# Importance Sampling with Unequal Support

Philip S. Thomas and Emma Brunskill
Carnegie Mellon University

## Abstract

Importance sampling is often used in machine learning when training and testing data come from different distributions. In this paper we propose a new variant of importance sampling that can reduce the variance of importance sampling-based estimates by orders of magnitude when the supports of the training and testing distributions differ. After motivating and presenting our new importance sampling estimator, we provide a detailed theoretical analysis that characterizes both its bias and variance relative to the ordinary importance sampling estimator (in various settings, which include cases where ordinary importance sampling is biased, while our new estimator is not, and *vice versa*). We conclude with an example of how our new importance sampling estimator can be used to improve estimates of how well a new treatment policy for diabetes will work for an individual, using only data from when the individual used a previous treatment policy.

## Introduction

A key challenge in artificial intelligence is to estimate the expectation of a random variable. Instances of this problem arise in areas ranging from planning and decision making (e.g., estimating the expected sum of rewards produced by a policy for decision making under uncertainty) to probabilistic inference. Although the estimation of an expected value is straightforward if we can generate many *independent and identically distributed* (i.i.d.) samples from the relevant probability distribution (which we refer to as the *target distribution*), we may not have generative access to the target distribution. Instead, we might only have data from a different distribution that we call the *sampling distribution*.

For example, in off-policy evaluation for reinforcement learning, the goal is to estimate the expected sum of rewards that a decision policy will produce, given only data gathered using some other policy. Similarly, in supervised learning, we may wish to predict the performance of a regressor or classifier if it were to be applied to data that comes from a distribution that differs from the distribution of the available data (e.g., we might predict the accuracy of a classifier for hand-written letters given that observed letter frequencies come from English, using a corpus of labeled letters collected from German documents).

More precisely, we consider the problem of estimating $\theta := \mathbf{E}[h(X)]$, where $h$ is a real-valued function and the expectation is over the random variable $X$, which is a sample from the target distribution. As input we assume access to $n$ i.i.d. samples from a sampling distribution that is different from the target distribution. A classical approach to this problem is to use *importance sampling* (IS), which reweighs the observed samples to account for the difference between the target and sampling distributions (Kahn, 1955). Importance sampling produces an unbiased but often high-variance estimate of $\theta$.

We introduce *importance sampling with unequal support* (US)—a simple new importance sampling estimator that can drastically reduce the variance of importance sampling when the supports of the sampling and target distributions differ. This setting with unequal support can occur, for example, in our earlier example where German documents might include symbols like ß, that the classifier will not encounter. US essentially performs importance sampling only on the data that falls within the support of the target distribution, and then scales this estimate by a constant that reflects the relative support of the target and sampling distributions.

US typically has lower variance than ordinary importance sampling (sometimes by orders of magnitude), and is unbiased in the important setting where at least one sample falls within the support of the target distribution. If no samples do, then none of the available data could have been generated by the target distribution, and so it is unclear what would make for a reasonable estimate. Furthermore, the conditionally unbiased nature of US is sufficient to allow for its use with concentration inequalities like Hoeffding's inequality to construct confidence bounds on $\theta$. By contrast, *weighted importance sampling* (Rubinstein, 1981) is another variant of importance sampling that can reduce variance, but which introduces bias that makes it incompatible with Hoeffding's inequality.

## Problem Setting and Importance Sampling

Let $f$ and $g$ be *probability density functions* (PDFs) for two distributions that we call the *target distribution* and *sampling distribution*, respectively. Let $h : \mathbb{R} \to \mathbb{R}$ be called the *evaluation function*. Let $\theta := \mathbf{E}_f[h(X)]$, where $\mathbf{E}_f$ denotes the expected value given that $f$ is the PDF of the random variable(s) in the expectation (in this case, just $X$). Let $F := \{x \in \mathbb{R} : f(x) \neq 0\}$, $G := \{x \in \mathbb{R} : g(x) \neq 0\}$, and $H := \{x \in \mathbb{R} : h(x) \neq 0\}$ be the supports of the target and sampling distributions, and the evaluation function, respectively. In this paper we will discuss techniques for estimating $\theta$ given $n \in \mathbb{N}_{>0}$ i.i.d. samples, $\mathbf{X}_n \coloneqq \{X_1, \ldots, X_n\}$, from the sampling distribution, and we focus on the setting where $F \cap H \subset G$—where the joint support of $F$ and $H$ is a *strict* subset of the support of $G$.

The importance sampling estimator,

$$\text{IS}(\mathbf{X}_n) \coloneqq t + \frac{1}{n} \sum_{i=1}^{n} \frac{f(X_i)}{g(X_i)}(h(X_i) - t), \qquad (1)$$

is a widely used estimator of $\theta$, where $t = 0$ (we consider non-zero values of $t$ later). If $F \cap H \subseteq G$, then $\text{IS}(\mathbf{X}_n)$ is a consistent and unbiased estimator of $\theta$. That is, $\text{IS}(\mathbf{X}_n) \xrightarrow{\text{a.s.}} \theta$ and $\mathbf{E}_g[\text{IS}(\mathbf{X}_n)] = \theta$ (we review this latter result in Property 1 in the supplemental document).

A *control variate* is a constant, $t \in \mathbb{R}$, that is subtracted from each $h(X_i)$ and then added back to the final estimate, as in (1) (Hammersley, 1960; Hammersley and Handscomb, 1964). Although control variates, $t(X_i)$, that depend on the sample, $X_i$, can be beneficial, for our later purposes we only consider constant control variates. Intuitively, including a constant control variate equates to estimating $\theta' \coloneqq \mathbf{E}_f[h'(X)]$ using importance sampling without a control variate, where $h'(x) = h(x) - t$, and then adding $t$ to the resulting estimate to get an estimate of $\theta$.

Later we show that the variance of importance sampling increases with $\theta^2$, and so applying importance sampling to $h$ results in higher variance than applying importance sampling to $h'$ with $t \approx \theta$, since then $\theta' \approx 0$. That is, by inducing a kind of normalization, a control variate can reduce the variance of estimates without introducing bias—a property that has made the inclusion of control variates a popular topic in some recent works using importance sampling (Dudík et al., 2011; Jiang and Li, 2016; Thomas and Brunskill, 2016). Although later we discuss control variates more, for simplicity our derivations focus on importance sampling estimators without control variates. There are also other extensions of the importance sampling estimator that can reduce variance—notably the weighted importance sampling estimator, which we compare to later, and which can provide large reductions of variance and mean squared error, but which introduces bias.

## An Illustrative Example

In this section we present an example that highlights the peculiar behavior of the IS estimator when $F \cap H \neq G$. Let $g(x) = 0.5$ if $x \in [0, 2]$ and $g(x) = 0$ otherwise, and let $f(x) = 1$ if $x \in [0, 1]$ and $f(x) = 0$ otherwise. So, $F = [0, 1]$ and $G = [0, 2]$. Let $h(x) = 1$ if $x \in [0, 1]$ and $h(x) = 0$ otherwise, so that $H = [0, 1]$. Notice that $\theta = 1$.

Since the sampling and target distributions are both uniform, an obvious estimator of $\theta$ (if $f$ and $g$ are known but $h$ is not) would be the average of the points that fall within $F$. Let $(\#X_i \in F)$ denote the number of samples in $\mathbf{X}_n$ that are in $F$. Formally, the obvious estimator is

$$\hat{\theta} \coloneqq \frac{1}{(\#X_i \in F)} \sum_{i=1}^{n} \mathbf{1}_F(X_i) h(X_i),$$

where $\mathbf{1}_A(x) = 1$ if $x \in \mathcal{A}$ and $\mathbf{1}_A(x) = 0$ otherwise. Given our knowledge of $h$, it is straightforward to show that this estimator is equal to 1 if $(\#X_i \in F) > 0$ and is undefined otherwise—it is exactly correct (has zero bias and variance) as long as at least one sample falls within $F$. If no samples fall within $F$, then we have only observed data that will never occur under the target distribution, and so we have no useful information about $\theta$. In this case, we might define our obvious estimator to return an arbitrary value, e.g., zero.

Perhaps surprisingly, the importance sampling estimator does not degenerate to this obvious estimator:

$$\text{IS}(\mathbf{X}_n) = \frac{1}{n} \sum_{i=1}^{n} \mathbf{1}_F(X_i) 2h(X_i) = \frac{2(\#X_i \in F)}{n}.$$

Since $\mathbf{E}_g[(\#X_i \in F)/n] = 1/2$, this estimate is correct in expectation, but does not have zero variance given that at least one sample falls within $F$. If more than $1/2$ of the samples fall within $F$, this estimate will be an over-estimate of $\theta$, and if fewer than $1/2$ of the samples fall within $F$, this estimate will be an under-estimate. Although correct on average, the importance sampling estimator has unnecessary additional variance relative to the obvious estimator.

## Importance Sampling with Unequal Support

We propose a new importance sampling estimator, *importance sampling with unequal support* (ISUS, or US for brevity), that *does* degenerate to the obvious estimator for our illustrative example. Intuitively, US prunes from $\mathbf{X}_n$ the samples that are outside $F$ (or more generally, outside some set $C$, that we define later) to construct a new data set, $\mathbf{X}'_n$, that has fewer samples. This new data set can be viewed as $(\#X_i \in F)$ i.i.d. samples from a different sampling distribution—a distribution with PDF $g'$, which is simply $g$, but truncated to only have support on $F$ and re-normalized to integrate to one. US then applies ordinary importance sampling to this new data set.

For generality, we allow US to prune from $\mathbf{X}_n$ all of the points that are not in a set, $C$, which can be defined many different ways, including $C \coloneqq F$ (as in our previous example). Our only requirement is that $F \cap H \subseteq C \subseteq G$. In order to compute US, we must compute a value,

$$c \coloneqq \int_C g(x) \, \mathrm{d}x,$$

which is the probability that a sample from the sampling distribution will be in $C$. In general, $C$ should be chosen to be as small as possible while still ensuring that both **1)** $F \cap H \subseteq C \subseteq G$ (so that informative samples are not discarded) and **2)** $c$ can be computed. Ideally, we would select $C = F \cap H$, however in some cases $c$ cannot be computed for this value of $C$. For example, in our later experiments we consider a problem where $h$ and $H$ are not known, but $F$ is, and so we can compute $c$ using $C = F$, but not $C = F \cap H$.

Let $k(\mathbf{X}_n) \coloneqq \sum_{i=1}^{n} \mathbf{1}_C(X_i)$ be the number of $X_i$ that are in $C$. The US estimator is then defined as:

$$\text{US}(\mathbf{X}_n) \coloneqq \frac{c}{k(\mathbf{X}_n)} \sum_{i=1}^{n} \frac{f(X_i)}{g(X_i)} h(X_i), \qquad (2)$$

if $k(\mathbf{X}_n) > 0$, and $\text{US}(\mathbf{X}_n) := 0$ if $k(\mathbf{X}_n) = 0$. This is equivalent to applying importance sampling to the pruned data set, $\mathbf{X}'_n$, since then $g'(x) = g(x)/c$ for $x \in C$. Also, in (2) we sum over all $n$ samples rather than just the $k(\mathbf{X}_n)$ samples in $C$ because $f(X_i)h(X_i) = 0$ for all $X_i$ not in $C$.

## Theoretical Analysis of US

We begin with two simple theorems that elucidate the relationship between IS and US. The proofs of both theorems are straightforward, but deferred to the supplemental document. First, Theorem 1 shows that, when $C = G$, US degenerates to IS. One case where $C = G$ is when the support of the target distribution and evaluation function are both equal to the support of the sampling distribution, i.e., when $F = H = G$, and so $C = G$ necessarily.

**Theorem 1.** *If $C = G$, then $\text{US}(\mathbf{X}_n) = \text{IS}(\mathbf{X}_n)$.*

Theorem 2 shows that, if we replace $c$ in the definition of US with an empirical estimate, $\hat{c}(\mathbf{X}_n) := k(\mathbf{X}_n)/n$, then US and IS are equivalent. This provides some intuition for why US tends to outperform IS when $C \subset G$—IS is US, but using an empirical estimate of $c$ (the probability that a sample falls within $C$), in place of its known value.

**Theorem 2.** *If we replace $c$ with an empirical estimate, $\hat{c}(\mathbf{X}_n) := k(\mathbf{X}_n)/n$, then $\text{US}(\mathbf{X}_n) = \text{IS}(\mathbf{X}_n)$.*

In Table 1 we summarize more theoretical results that clarify the differences between IS and US in several settings. The first setting (denoted by a † in Table 1) is the standard setting where we consider the ordinary expected value and variance of the two estimators. The second setting (denoted by a ‡ in Table 1) conditions on the event that at least one sample falls within $C$, that is, the event that $k(\mathbf{X}_n) > 0$. This is a reasonable setting to consider if one takes the view that no estimate should be returned if all of the samples are outside $C$. That is, if the pruned data set, $\mathbf{X}'_n$, is empty, then no estimate should be produced or considered (just as IS does not produce an estimate when $n = 0$—when there are no samples at all). Finally, the third setting (denoted by a ⋆ in Table 1) conditions on the event that $k(\mathbf{X}_n) = \kappa$—that a specific constant number of the $n$ samples are in $C$.

Table 1 and the theorems that it references use additional symbols that we review here. Let $\rho := \Pr(k(\mathbf{X}_n) > 0) = 1 - (1-c)^n$ be the probability that at least one of $n$ samples is in $C$. Let $\text{Var}_g(\cdot)$ denote the variance given that the random variables within the parenthesis are sampled from the distribution with PDF $g$. Let

$$v := \text{Var}_g\left(\frac{f(X)}{g(X)}h(X)\bigg|X \in C\right)$$

be the conditional variance of the importance sampling estimate when using a single sample and given that the sample is in $C$. Let $B(n,c)$ denote the binomial distribution with parameters $n$ and $c$ and let $\mathbf{E}_{B(n,c)}$ denote the expected value given that $\kappa \sim B(n,c)$.

Although the proofs of the claims in Table 1 are some of the primary contributions of this work, we defer them to the supplemental document because they are straightforward (though lengthy) and do not provide further insights into the results. The primary result of Table 1 is that US is unbiased and often has lower variance in the key setting of interest: when at least one sample is in the support of the target distribution—when $k(\mathbf{X}_n) > 0$. We find this setting compelling because, when no samples are in $F$, little can be inferred about $\mathbf{E}_f[h(X)]$.

In this setting (denoted by ‡ in Table 1) US is an unbiased estimator, while IS is not (although the bias of IS does go to zero as $n \to \infty$).[1] To understand the source of this bias, consider the bias of IS given that $k(\mathbf{X}_n) = \kappa$—the ⋆ setting in Table 1. In this case, $\mathbf{E}_g[\text{IS}(\mathbf{X}_n)] = \frac{\kappa}{cn}\theta$. Recall that IS uses an empirical estimate of $c$, i.e., $\hat{c} \approx \frac{\kappa}{n}$ (as discussed in Theorem 2). When this estimate is correct, terms in $\frac{\kappa}{cn}\theta$ cancel, making IS unbiased. Thus, the bias of IS when conditioning on the event that $k(\mathbf{X}_n) > 0$ stems from IS's use of an estimate of $c$.

Next we discuss the variance of the two estimators given that at least one sample falls within $C$, i.e., in the ‡ setting. First consider how the variances of IS and US change as $c \to 0$—that is, as the differences between the supports of the sampling and target distributions increases. Specifically, let $c_i := \frac{1}{i}$ for $i \in \mathbb{N}_{>0}$. We then have that: $\text{Var}(\text{IS}(\mathbf{X}_n)|k(\mathbf{X}_n) > 0, c_i) \geq \frac{c_i v}{n\rho} = \frac{v}{n\rho i} \geq \frac{v}{ni}$, since $\rho \in (0,1]$, and $\text{Var}(\text{US}(\mathbf{X}_n)|k(\mathbf{X}_n) > 0, c_i) = (v/i^2)\mathbf{E}_{B(n,c)}[1/\kappa|\kappa > 0] \leq v/i^2$, since $\mathbf{E}_{B(n,c)}[\kappa^{-1}|\kappa > 0] \leq 1$. Thus, as $i \to \infty$ (as $c \to 0$ logarithmically), and given some fixed $n$ and $v$, the variance of US goes to zero much faster than the variance of IS. The variance of US (as a function of $i$) converges to zero linearly (or faster) with a rate of at most 1 while the variance of IS converges to zero sublinearly (at best, logarithmically).

Next note that the variance of US in this setting is independent of $\theta^2$, but the variance of IS increases with $\theta^2$ (see Property 3 in the supplemental document, applied to Theorem 9). To ameliorate this issue, a control variate, $t$, can be used to center the data so that $\theta \approx 0$. However, since $\theta$ is not known *a priori*, selecting $t = \theta$ is not practical. The term that scales with $\theta^2$ in the variance of IS given that $k(\mathbf{X}_n) > 0$ therefore means that the variance of IS depends on the quality of the control variate—poor control variates can cause IS to have high variance. By contrast, the variance of US in this setting does not have a term that scales with $\theta^2$, and so the quality of the control variate is less important.[2]

There is a rare case when IS can have a lower variance than US. First, we assume that the control variate is perfect so that $\theta = 0$ (which, as discussed before, is impractical) and consider the term that scales with $v$. From this term, it is clear that US will have lower variance that IS if:

$$c^2 \mathbf{E}_{B(n,c)}[\kappa^{-1}|\kappa > 0] \leq \frac{c}{n\rho}. \quad (3)$$

---

[1] If we do not condition on the event that $k(\mathbf{X}_n) > 0$, then US is a *biased* estimator of $\theta$. This is because it is unclear how to define $\text{US}(\mathbf{X}_n)$ when $k(\mathbf{X}_n) = 0$, and we chose (arbitrarily) to define it to be 0. However, the bias of $\text{IS}(\mathbf{X}_n)$ in this setting converges quickly to zero, since $\rho$ (the probability that no samples fall within $C$) converges quickly to one as $n \to \infty$.

[2] The quality of the control variate can still impact the variance of estimates though, since it can change $v$.

| | $\mathbf{E}_g[\cdot]$ † | $\mathbf{E}_g[\cdot]$ ‡ | $\mathbf{E}_g[\cdot]$ ⋆ | **Variance**† | **Variance**‡ | **Strongly Consistent** |
|---|---|---|---|---|---|---|
| IS | $\theta$ (Property 1) | $\frac{1}{\rho}\theta$ (Theorem 6) | $\frac{\kappa}{cn}\theta$ (Theorem 5) | $\frac{1}{n}\left(cv + \theta^2\left(\frac{1}{c}-1\right)\right)$ (Theorem 11) | $v\frac{c}{n\rho} + \theta^2\frac{c\rho(n-1)+\rho-cn}{cn\rho^2}$ (Theorem 9) | Yes († and ‡) |
| US | $\rho\theta$ (Theorem 7) | $\theta$ (Theorem 4) | $\theta$ (Theorem 3) | $\rho c^2 v \mathbf{E}_{B(n,c)}[\kappa^{-1}\|\kappa>0]$ $+\theta^2\rho(1-\rho)$ (Theorem 10) | $c^2 v \mathbf{E}_{B(n,c)}[\kappa^{-1}\|\kappa>0]$ (Theorem 8) | Yes († and ‡) |

Table 1: Theoretical properties of IS and US estimators. † = given no conditions. ‡ = conditioned on the event that $k(\mathbf{X}_n) > 0$—that at least one sample is in $C$. ⋆ = conditioned on the event that $k(\mathbf{X}_n) = \kappa$—that exactly $\kappa$ of $n$ samples are in $C$. All theorems require the assumption that $F \cap H \subseteq G$. The consistency results follow immediately from the fact that the biases and variances all converge to zero as $n \to \infty$ (Thomas and Brunskill, 2016, Lemma 3).

Notice that this inequality depends only on $n$ and $c$, which must both be known in order to implement US, and so we can test *a priori* whether US will have lower variance than IS. That is, if (3) holds, then US will have lower variance than IS, given that $k(\mathbf{X}_n) > 0$. However, if (3) does not hold, it does not mean that IS will have lower variance than US unless the perfect (typically unknown) control variate is used so that $\theta = 0$.

### Application to Illustrative Example

Because neither method is always superior, here we consider the application of IS and US to the illustrative example to see when each method works best, and by how much. We consider the setting where $C = F$, but modify the example slightly. First, although the target distribution is always uniform, we allow for its support to be scaled. Specifically, we define the support of $f$ to be $[0, F_{\max}]$, where $F_{\max} \in (0, 2]$. When $F_{\max}$ is small, it corresponds to significant differences in support, while large $F_{\max}$ correspond to small differences (when $F_{\max} = 2$, $C = F = G$ and so the two estimators are equivalent). We also modify $h$ to allow for various values of $\theta$. Specifically, we define $h(x) = -1 + \theta$ if $x < F_{\max}/2$ and $h(x) = 1 + \theta$ if $x \geq F_{\max}/2$. Notice that, although we defined $h$ in terms of $\theta$, $\theta$ remains $\mathbf{E}_f[h(X)]$, and also that using this definition of $h$ and $\theta = 0$ is an instance that is particularly favorable to IS.

For this example, it is straightforward to verify that $v = 4/F_{\max}^2$ for any definition of $\theta$, and $c = F_{\max}/2$. Given these two values (and $\theta$), we can compute the bias and variance of each estimator. The biases and variances of the two estimators for various settings are depicted in Figure 1. Notice that US is always competitive with IS, although the reverse is not true. Particularly, when $F_{\max}$ is small (so that $c$ is small), or when $\theta$ is large, US can have orders of magnitude lower variance than IS. Also, as $n$ increases, the two estimators become increasingly similar, since the empirical estimate of $c$ used by IS becomes increasingly accurate, although US is still vastly superior to IS even when $n$ is large if $c$ is correspondingly small. This matches our theoretical analysis from the previous section: we expect US to perform better when $c$ is small (by our convergence rate analysis) or when $\theta^2$ is large (due to US's lesser dependence on the quality of the control variate), and we expect the two estimators to becomes increasingly similar as $n \to \infty$ (because $\hat c$ becomes increasingly similar to $c$).

Notice also that gains are not only obtained when $c$ is so small relative to $n$ that no samples are expected to fall within $C$ (a relatively uninteresting setting). For example, the right-most plot in Figure 1 shows that with $F_{\max} = 0.5$, where $\Pr(k(\mathbf{X}_n) > 0) = \rho = 1 - \frac{1}{2^{50}} \approx 1$, the MSE of US is approximately 0.086, while the MSE of IS is approximately 6.08—US is has roughly 1/70 the MSE of IS (1/8 the RMSE).

Perhaps surprisingly, there are cases where IS has lower variance than US (even when both are unbiased, since $\theta = 0$). For example, consider the plot with $\theta = 0$ and $n = 10$, and the position on the horizontal axis that corresponds to $F_{\max} = 1.0$. This is one case where IS is marginally better than US (it has lower variance in both settings, and neither estimator is biased). Intuitively, the IS estimator includes the points outside the support of $F$, although they have associated values, $h(X_i) = 0$, which pulls the importance sampling estimate towards zero. In this case, when $\theta = 0$, this extra pull towards zero happens to be beneficial. However, to remain unbiased given the pull towards zero, IS also increases the magnitudes of the weights associated with points in $F$, which incurs additional variance. When $F_{\max}$ is small enough, this additional variance outweighs the variance reduction that results from the extra pull towards zero, and so US is again superior. This intuition is supported by the fact that in Figure 1 IS does not outperform US for small $F_{\max}$ or $\theta \geq 1$, since then a pull towards zero is detrimental.

Finally, we consider the use of IS and US to create high-confidence upper and lower bounds on $\theta$ using a concentration inequality (Massart, 2007) like Hoeffding's inequality (Hoeffding, 1963). If $b$ denotes the range of the function $f(x)h(x)/g(x)$, for $x \in G$, then using Hoeffding's inequality, we have that $\text{IS}(\mathbf{X}_n) - b\sqrt{\ln(1/\delta)/(2n)}$ is a $1 - \delta$ confidence lower bound on $\theta$. Similarly, we can use US with Hoeffding's inequality to create a $1 - \delta$ confidence lower bound: $\text{US}(\mathbf{X}_n) - cb\sqrt{\ln(1/\delta)/(2k(\mathbf{X}_n))}$, since the range of the $k(\mathbf{X}_n)$ i.i.d. random variables averaged by $\text{US}(\mathbf{X}_n)$ is $cb$. Notice that, if $k(\mathbf{X}_n) = 0$, then this second estimator is undefined (one might define the lower bound to be a known lower bound on $\theta$ in this setting). Although we expect that $k(\mathbf{X}_n) \approx cn$, the resulting $c$ in the denominator of the US-based bound is within the square root, while the $c$ in the numerator is not, and so the bound constructed using US should tend to be tighter when $c$ is small.

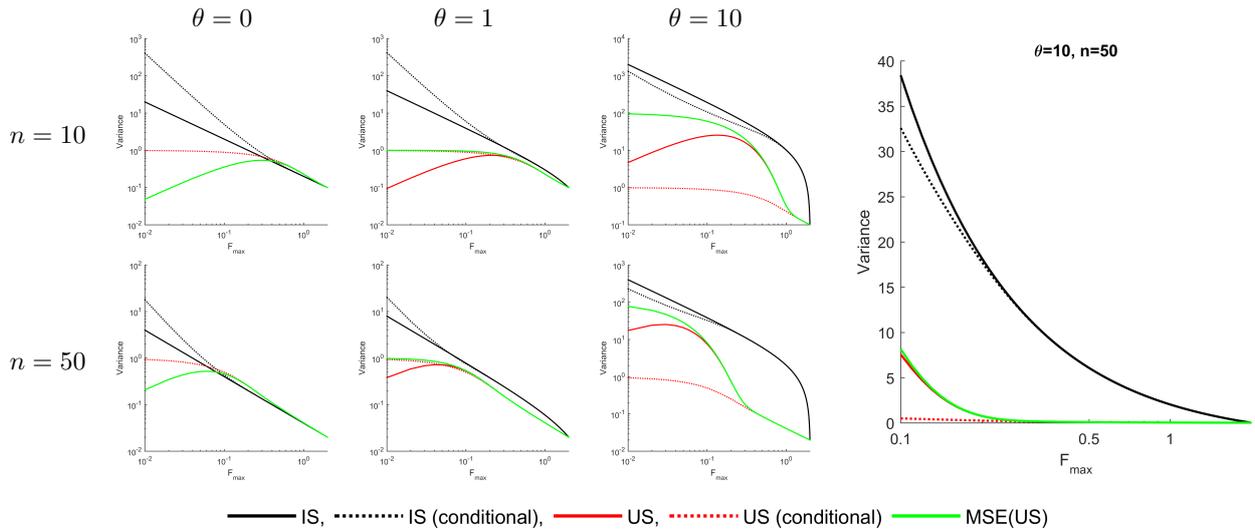

Figure 1: The variances of IS and US across various settings of $n$ and $\theta$ (denoted along the left and top). At a glance, notice that the red and green curves (US) tend to be below the black curves (IS), particularly when considering the logarithmic scale of the vertical axes. The dotted lines show the variance conditioned on the event that $k(\mathbf{X}_n) > 0$. The green line shows the mean squared error of the US estimator (without any conditions), which shows that the variance reduction of US is not completely offset by increased bias (compare the solid black and green curves). When $\theta = 0$ the green line obscures the solid red line. The plot on the right shows a zoomed-in view of the $\theta = 10, n = 50$ plot without the logarithmic vertical axis.

## Application to Diabetes Treatment

We applied US and IS to the problem of predicting the effectiveness of altering the treatment policy for diabetes 1 for a particular individual. That is, we would like to use prior data from when the individual was treated with one treatment policy to estimate how well a related policy would work. The treatment policy is parameterized by two numbers, CR and CF, and dictates how much insulin a person should inject prior to eating a meal in order to keep his or her blood glucose close to optimum levels. CR and CF are typically specified by a diabetologist and tweaked during follow-up visits every 3–6 months. If follow-up visits are not an option, recent research has suggested using reinforcement learning algorithms to tune CR and CF (Bastani, 2014).

Here we focus on a sub-problem of improving CR and CF—using data collected from an initial range of admissible values of CR and CF to predict how well a new range of values for CR and CF would perform. When collecting data, CR and CF are drawn uniformly from an initial admissible range, and then used for one day (which we view as one episode of a Markov decision process). The performance during each day is measured using an objective function similar to the reward function proposed by Bastani (2014), which measures the deviation of blood glucose from optimum levels, with larger penalties for low blood glucose levels. We refer to the measure of how good the outcome was from one day as the *return* associated with that day, with larger values being better. Using approximately 30 days of data, our goal is to estimate the expected return if a different distribution of CR and CF were to be used.

We consider a specific *in silico* person—a person simulated using a metabolic simulator. We used the subject "Adult#003" in the Type 1 Diabetes Metabolic Simulator (T1DMS) (Dalla Man et al., 2014)—a simulator that has been approved by the US Food and Drug Administration as a substitute for animal trials in pre-clinical testing of treatment policies for type 1 diabetes. During each day, the subject is given three or four meals of randomized sizes at randomized times, similar to the experimental setup proposed by Bastani (2014). As a result of this randomness, and the stochastic nature of the T1DMS model, applying the same values of CR and CF can produce different returns if used for multiple days. After analyzing the performance of many CR and CF pairs, we selected an initial range that results in good performance: CR $\in [8.5, 11]$ and CF $\in [10, 15]$. Using a large number of samples, we computed an estimate of the expected return if different CR and CF values are used for a single day—this estimate is depicted in Figure 2.

As described by Bastani (2014), when the value of CR is set appropriately, performance is robust to changes in CF. We therefore focus on possible changes to CR. Specifically, we consider new treatment policies where CF remains sampled from the uniform distribution over $[10, 15]$, but where CR is sampled from the truncated normal distribution over $[\text{CR}_{\min}, 11]$, with mean 11 and standard deviation $11 - \text{CR}_{\min}$. This distribution places the largest probability densities at the upper end of the range of CR, which favors better policies. As $\text{CR}_{\min}$ increases towards 11, the support of the sampling distribution and target distribution become increasingly different ($c = (11 - \text{CR}_{\min})/2.5$) and the expected return increases.

For each value of $\text{CR}_{\min}$ (each of which corresponds to a value of $c$), we performed 2,433 trials, each of which involved sampling 30 days of data from the sampling distribution and then using IS, US, and *weighted importance sampling* (WIS) to estimate the expected return if CR and CF were sampled from the target distribution. Figure 2 displays the bias, variance and *mean squared error* (MSE) of these 2,433 estimates, using an estimate of ground truth computed using Monte Carlo sampling. Figure 2 also shows the impact of providing a constant control variate to all the estimators: the chosen control variate was the expected return under the sampling distribution.

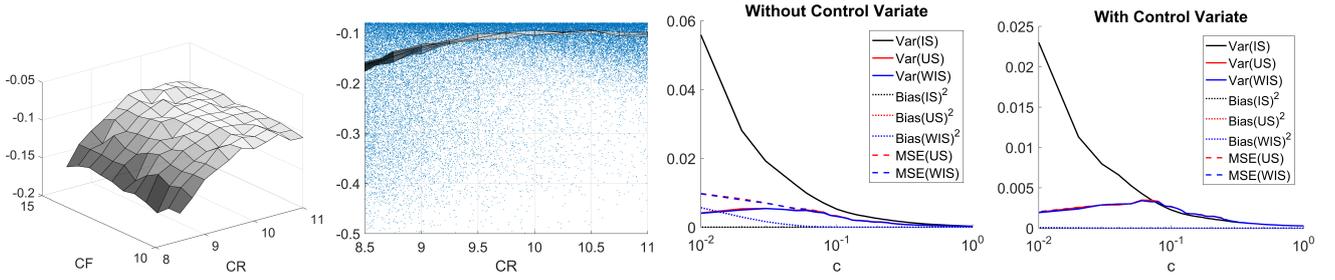

Figure 2: The first and second plots show an estimate of the expected return for various CR and CF, from two different angles. The second plot includes points depicting the returns observed from using different values of CR and CF for a day—notice the high variance. The two plots on the right depict the bias, variance, and MSE of IS, US, and WIS (without any conditioning) for various values of $c$ and both without (third plot) and with (fourth plot) a control variate. The curves for US are largely obscured by the corresponding curves for WIS. Notice that the variance of IS approaches 0.06, which is enormous given that the difference between the best and worst CR and CF pairs possible under the sampling policy is approximately 0.06.

Notice that we see the same trend as in the illustrative example—for small $c$ (the best treatment policies, which have small ranges of CR), US significantly outperforms IS. Furthermore, when a decent control variate is not used, the benefits of US are increased, even when controlling for the resulting bias by measuring the mean squared error. We also computed the biases and variances given that $k(\mathbf{X}_n) > 0$, and observed similar results (not shown), which favored US slightly more. Notice that WIS and US perform very similarly. Indeed, if the sampling and target distributions are both uniform, it is straightforward to verify that WIS and US are equivalent. In other experiments (not shown) we found that WIS yields lower variance than US when the target distribution is modified to be even less like the uniform distribution.

However, it is often important to be able to produce confidence intervals around estimates (especially when data is limited), and since WIS is biased, it cannot be used with standard concentration inequalities. We used Hoeffding's inequality to compute a 90% confidence interval around the estimates produced by IS and US (without control variates and with $\text{CR}_{\min} = 10.375$, so that $c = 1/4$) using various numbers of samples (days of data). The mean confidence intervals are depicted in Figure 3, which also shows a Monte Carlo estimate of $\theta$, as well as deterministic domain-specific upper and lower bounds on $h(X)$ (denoted by "$h$ range" in the legend). If $k(\mathbf{X}_n) = 0$, then US is not defined, and so the confidence intervals shown for US are averaged only over the instances where $k(\mathbf{X}_n) > 0$. To show how often US returns a solution, Figure 3 also shows $\rho$—the probability that US will produce a confidence bound—using the right vertical axis for scale.

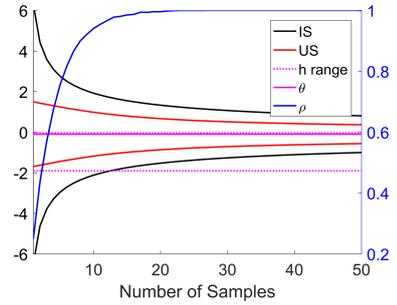

Figure 3: Confidence bounds using IS and US.

US produces a much tighter confidence interval than IS in all cases. Furthermore, the setting where US often does not return a bound corresponds to the setting where IS produces a confidence interval that is outside the deterministic bound on $h(X)$—a trivial confidence interval. In additional experiments (not shown) we defined the bounds to be truncated to always be within the deterministic bounds on $h(X)$ and define the bound produced using US to be conservative (equal to the deterministic bounds) when $k(\mathbf{X}_n) = 0$. In this experiment we saw similar results—the confidence intervals produced using US were much tighter than those using IS.

## Conclusion and Future Work

We have presented a simple new variant of importance sampling, US. Our analytical and empirical results suggest that US can significantly outperform ordinary importance sampling, and we provide an *a priori* calculation to check for the rare cases where it can perform slightly worse. Unlike some other IS estimators that have been developed to reduce variance (like WIS), US is unbiased given mild conditions that still permit the easy computation of confidence intervals.

# Supplemental Document

In this supplemental document we prove the various properties and theorems referenced earlier (particularly those in Table 1).

**Property 1.** *If $F \cap H \subseteq G$ then $\mathbf{E}_g[\text{IS}(\mathbf{X}_n)] = \theta$.*

*Proof.*

$$\mathbf{E}_g[\text{IS}(\mathbf{X}_n)] \stackrel{(a)}{=} \mathbf{E}_g\left[\frac{f(X)}{g(X)}h(X)\right] = \int_G g(x)\frac{f(x)}{g(x)}h(x)\,\mathrm{d}x$$

$$\stackrel{(b)}{=} \int_{F \cap H} f(x)h(x)\,\mathrm{d}x = \mathbf{E}_f[h(X)] = \theta,$$

where **(a)** holds because $\text{IS}(\mathbf{X}_n)$ is the mean of $n$ independent and identically distributed random variables, and **(b)** holds because $\forall x \in G \setminus (F \cap H), f(x) = 0$. ∎

We now provide a proof of Theorem 1, which states that if $C = G$, then $\text{US}(\mathbf{X}_n) = \text{IS}(\mathbf{X}_n)$.

*Proof.* In this setting, $c = \int_G g(x)\,\mathrm{d}x = 1$ and since every $X_i$ must be within $C$, $k(\mathbf{X}_n) = n$. So,

$$\text{US}(\mathbf{X}_n) = \frac{c}{k(\mathbf{X}_n)}\sum_{i=1}^n \frac{f(X_i)}{g(X_i)}h(X_i)$$

$$= \frac{1}{n}\sum_{i=1}^n \frac{f(X_i)}{g(X_i)}h(X_i). \qquad \blacksquare$$

We now provide a proof of Theorem 2, which states that if we replace $c$ with an empirical estimate, $\hat{c}(\mathbf{X}_n) := n^{-1}k(\mathbf{X}_n)$, then $\text{US}(\mathbf{X}_n) = \text{IS}(\mathbf{X}_n)$.

*Proof.* Using the empirical estimate, $\hat{c}(\mathbf{X}_n)$, in place of $c$ within US we have:

$$\text{US}(\mathbf{X}_n) = \frac{\hat{c}(\mathbf{X}_n)}{k(\mathbf{X}_n)}\sum_{i=1}^n \frac{f(X_i)}{g(X_i)}h(X_i)$$

$$= \frac{k(\mathbf{X}_n)}{nk(\mathbf{X}_n)}\sum_{i=1}^n \frac{f(X_i)}{g(X_i)}h(X_i)$$

$$= \frac{1}{n}\sum_{i=1}^n \frac{f(X_i)}{g(X_i)}h(X_i)$$

$$= \text{IS}(\mathbf{X}_n). \qquad \blacksquare$$

**Theorem 3.** *If $F \cap H \subseteq G$ and $\kappa \in \mathbb{N}_{>0}$, then*

$$\mathbf{E}_g[\text{US}(\mathbf{X}_n)|k(\mathbf{X}_n) = \kappa] = \theta.$$

*Proof.* Let $\Pr_g(X \in C)$ denote the probability that a sample, $X$, from the sampling distribution is in $C$.

$$\mathbf{E}_g[\text{US}(\mathbf{X}_n)|k(\mathbf{X}_n) = \kappa]$$

$$= \mathbf{E}_g\left[\frac{c}{\kappa}\sum_{i=1}^n \frac{f(X_i)}{g(X_i)}h(X_i)\bigg|k(\mathbf{X}_n) = \kappa\right]$$

$$\stackrel{(a)}{=} \mathbf{E}_g\left[\frac{c}{\kappa}\sum_{i=1}^\kappa \frac{f(X_i)}{g(X_i)}h(X_i)\bigg|\forall i \in \{1,\ldots,\kappa\}, X_i \in C\right]$$

$$\stackrel{(b)}{=} \mathbf{E}_g\left[c\frac{f(X)}{g(X)}h(X)\bigg|X \in C\right]$$

$$\stackrel{(c)}{=} \int_C \frac{g(x)}{\Pr_g(X \in C)}c\frac{f(x)}{g(x)}h(x)\,\mathrm{d}x$$

$$\stackrel{(d)}{=} \int_C \frac{g(x)}{c}c\frac{f(x)}{g(x)}h(x)\,\mathrm{d}x$$

$$= \int_C f(x)h(x)\,\mathrm{d}x$$

$$\stackrel{(e)}{=} \mathbf{E}_f[h(X)],$$

where **(a)** holds because $f(X_i) = 0$ for all but $\kappa$ of the terms in the summation, and so (by re-ordering the $X_i$ so that these $\kappa$ terms have indices $1,\ldots,\kappa$) we need only sum to $\kappa$ rather than $n$, **(b)** holds because the summation is over $\kappa$ independent and identically distributed random variables, **(c)** holds by the definition of conditional expectations, **(d)** holds because $\Pr_g(X \in C) = c$, and **(e)** holds because $F \cap H \subseteq C$. ∎

**Theorem 4.** *If $F \cap H \subseteq G$ then*

$$\mathbf{E}_g[\text{US}(\mathbf{X}_n)|k(\mathbf{X}_n) > 0] = \theta.$$

*Proof.*

$$\mathbf{E}_g[\text{US}(\mathbf{X}_n)|k(\mathbf{X}_n) > 0]$$

$$= \sum_{\kappa=1}^n \frac{\Pr(k(\mathbf{X}_n) = \kappa | k(\mathbf{X}_n) > 0)}{\Pr(k(\mathbf{X}_n) > 0)}\mathbf{E}_g[\text{US}(\mathbf{X}_n)|k(\mathbf{X}_n) = \kappa]$$

$$\stackrel{(a)}{=} \sum_{\kappa=1}^n \frac{\Pr(k(\mathbf{X}_n) = \kappa | k(\mathbf{X}_n) > 0)}{\Pr(k(\mathbf{X}_n) > 0)}\theta$$

$$= \theta \sum_{\kappa=1}^n \frac{\Pr(k(\mathbf{X}_n) = \kappa | k(\mathbf{X}_n) > 0)}{\Pr(k(\mathbf{X}_n) > 0)}$$

$$= \theta,$$

where **(a)** holds because, by Theorem 3, $\mathbf{E}[\text{US}(\mathbf{X}_n)|k(\mathbf{X}_n) = \kappa] = \theta$. ∎

**Theorem 5.** *If $F \cap H \subseteq G$ and $\kappa \in \mathbb{N}_{>0}$, then*

$$\mathbf{E}_g[\text{IS}(\mathbf{X}_n)|k(\mathbf{X}_n) = \kappa] - \theta = \left(\frac{\kappa}{cn} - 1\right)\theta. \qquad (4)$$

*Proof.* Following roughly the same steps as used to prove

Theorem 3 we have that:

$$\begin{aligned}
&\mathbf{E}_g[\,\mathrm{IS}(\mathbf{X}_n)|k(\mathbf{X}_n) = \kappa]\\
&=\mathbf{E}_g\left[\frac{1}{n}\sum_{i=1}^{n}\frac{f(X_i)}{g(X_i)}h(X_i)\bigg|k(\mathbf{X}_n) = \kappa\right]\\
&=\mathbf{E}_g\left[\frac{1}{n}\sum_{i=1}^{\kappa}\frac{f(X_i)}{g(X_i)}h(X_i)\bigg|\forall i \in \{1,\ldots,\kappa\}, X_i \in C\right]\\
&=\mathbf{E}_g\left[\frac{\kappa}{n}\frac{f(X_1)}{g(X_1)}h(X_1)\bigg|X_1 \in C\right]\\
&=\int_C \frac{g(x)}{c}\frac{\kappa}{n}\frac{f(x)}{g(x)}h(x)\,\mathrm{d}x\\
&=\frac{\kappa}{cn}\mathbf{E}_f[h(X)]\\
&=\frac{\kappa}{cn}\theta,
\end{aligned}$$

and so (4) follows. ∎

**Theorem 6.** *If $F \cap H \subseteq G$ then*

$$\mathbf{E}_g[\mathrm{IS}(\mathbf{X}_n)|k(\mathbf{X}_n) > 0] = \frac{1}{1 - (1-c)^n}\theta.$$

*Proof.* Recall from Property 1 that $\mathbf{E}_g[\mathrm{IS}(\mathbf{X}_n)] = \theta$. By marginalizing over whether or not $k(\mathbf{X}_n) > 0$, we also have that:

$$\begin{aligned}
\mathbf{E}_g[\mathrm{IS}(\mathbf{X}_n)] =& \Pr(k(\mathbf{X}_n) > 0)\mathbf{E}_g[\mathrm{IS}(\mathbf{X}_n)|k(\mathbf{X}_n) > 0]\\
&+ \Pr(k(\mathbf{X}_n) = 0)\mathbf{E}_g[\mathrm{IS}(\mathbf{X}_n)|k(\mathbf{X}_n) = 0].
\end{aligned}$$

So,

$$\begin{aligned}
&\mathbf{E}_g[\,\mathrm{IS}(\mathbf{X}_n)|k(\mathbf{X}_n) > 0]\\
&=\frac{\theta - \Pr(k(\mathbf{X}_n) = 0)\mathbf{E}_g[\mathrm{IS}(\mathbf{X}_n)|k(\mathbf{X}_n) = 0]}{\Pr(k(\mathbf{X}_n) > 0)}\\
&\stackrel{(a)}{=}\frac{\theta}{1 - (1-c)^n},
\end{aligned}$$

where **(a)** holds because $\mathbf{E}_g[\mathrm{IS}(\mathbf{X}_n)|k(\mathbf{X}_n) = 0] = 0$ and $\Pr(k(\mathbf{X}_n) > 0) = 1 - \Pr(k(\mathbf{X}_n) = 0) = 1 - (1-c)^n$. ∎

**Theorem 7.** *If $F \cap H \subseteq G$, then*

$$\mathbf{E}_g[\mathrm{US}(\mathbf{X}_n)] = (1 - (1-c)^n)\theta.$$

*Proof.*

$$\begin{aligned}
&\mathbf{E}_g[\,\mathrm{US}(\mathbf{X}_n)]\\
&=\underbrace{\Pr(k(\mathbf{X}_n) > 0)}_{=1-(1-c)^n}\underbrace{\mathbf{E}_g[\mathrm{US}(\mathbf{X}_n)|k(\mathbf{X}_n) > 0]}_{=\theta,\text{ by Theorem 4}}\\
&\quad + \Pr(k(\mathbf{X}_n) = 0)\underbrace{\mathbf{E}_g[\mathrm{US}(\mathbf{X}_n)|k(\mathbf{X}_n) = 0]}_{=0}\\
&=(1 - (1-c)^n)\theta.
\end{aligned}$$
∎

Before continuing, recall the following property (which we prove for completeness):

**Property 2.** *Let $X_1,\ldots,X_n$ be $n$ independent and identically distributed random variables, each with finite mean and variance. Then,*

$$\mathbf{E}\left[\left(\frac{1}{n}\sum_{i=1}^{n}X_i\right)^2\right] = \frac{1}{n}\mathrm{Var}(X_1) + \mathbf{E}[X_1]^2.$$

*Proof.* Recall that

$$\mathrm{Var}\left(\frac{1}{n}\sum_{i=1}^{n}X_i\right) = \mathbf{E}\left[\left(\frac{1}{n}\sum_{i=1}^{n}X_i\right)^2\right] - \mathbf{E}\left[\frac{1}{n}\sum_{i=1}^{n}X_i\right]^2.$$

So, by rearranging terms:

$$\mathbf{E}\left[\left(\frac{1}{n}\sum_{i=1}^{n}X_i\right)^2\right] = \frac{1}{n^2}\mathrm{Var}\left(\sum_{i=1}^{n}X_i\right) + \frac{1}{n^2}\mathbf{E}\left[\sum_{i=1}^{n}X_i\right]^2.$$

Since the $X_i$ are independent and identically distributed, we therefore have that:

$$\begin{aligned}
\mathbf{E}\left[\left(\frac{1}{n}\sum_{i=1}^{n}X_i\right)^2\right] &= \frac{1}{n^2}n\,\mathrm{Var}(X_1) + \frac{1}{n^2}n^2\mathbf{E}[X_1]^2\\
&= \frac{1}{n}\mathrm{Var}(X_1) + \mathbf{E}[X_1]^2.
\end{aligned}$$
∎

**Theorem 8.** *If $F \cap H \subseteq G$ then*

$$\mathrm{Var}_g(\mathrm{US}(\mathbf{X}_n)|k(\mathbf{X}_n > 0)) = c^2 v \mathbf{E}_{B(n,c)}\left[\frac{1}{\kappa}\bigg|\kappa > 0\right].$$

*Proof.*

$$\begin{aligned}
&\mathrm{Var}_g(\mathrm{US}(\mathbf{X}_n)|k(\mathbf{X}_n) > 0)\\
&=\mathbf{E}_g[\mathrm{US}(\mathbf{X}_n)^2|k(\mathbf{X}_n) > 0] - \mathbf{E}_g[\mathrm{US}(\mathbf{X}_n)|k(\mathbf{X}_n) > 0]^2\\
&=\mathbf{E}_g[\mathrm{US}(\mathbf{X}_n)^2|k(\mathbf{X}_n) > 0] - \theta^2\\
&=\left(\sum_{\kappa=1}^{n}\frac{\Pr(k(\mathbf{X}_n) = \kappa)}{\Pr(k(\mathbf{X}_n) > 0)}\mathbf{E}_g[\mathrm{US}(\mathbf{X}_n)^2|k(\mathbf{X}_n) = \kappa]\right) - \theta^2.
\end{aligned}$$
(5)

We will write $\mathbf{y}$ to denote a vector in $\mathbb{R}^n$, the elements of which are $y_1,\ldots,y_n \in \mathbb{R}$. We also write $\mathbf{y}_{i:j}$ to denote the $i^{\mathrm{th}}$ through $j^{\mathrm{th}}$ entries of $\mathbf{y}$, i.e., $\mathbf{y}_{i:j} := [y_i, y_{i+1},\ldots,y_{j-1}, y_j]$. Let $G_\kappa^n = \{\mathbf{y} \in G^n : k(\mathbf{y}) = \kappa\}$ be the set of all possible tuples of $n$ samples where exactly $\kappa$ are in $C$. We also overload the definition of $g$ by defining $g(\mathbf{y}) := \prod_{i=1}^{n}g(y_i)$. Using this notation, we have that (where $\ldots$ are used to denote that a long line is split across multiple lines via scalar multiplication):

$$\mathbf{E}_g[\,\mathrm{US}(\mathbf{X}_n)^2|k(\mathbf{X}_n)=\kappa]$$
$$=\int_{G_\kappa^n}\frac{g(\mathbf{y})}{\Pr(k(\mathbf{X}_n)=\kappa)}\,\mathrm{US}(\mathbf{y})^2\,\mathrm{d}\mathbf{y}$$
$$\stackrel{(a)}{=}\frac{\binom{n}{\kappa}}{\Pr(k(\mathbf{X}_n)=\kappa)}\int_{C^\kappa}\int_{(G\setminus C)^{n-\kappa}}g(\mathbf{y})\,\mathrm{US}(\mathbf{y})^2\,\mathrm{d}\mathbf{y}_{1:\kappa}\,\mathrm{d}\mathbf{y}_{\kappa+1:n}$$
$$\stackrel{(b)}{=}\frac{\binom{n}{\kappa}}{\Pr(k(\mathbf{X}_n)=\kappa)}\int_{C^\kappa}\int_{(G\setminus C)^{n-\kappa}}g(\mathbf{y}_{1:\kappa})g(\mathbf{y}_{\kappa+1:n})\cdots$$
$$\mathrm{US}(\mathbf{y}_{1:\kappa})^2\,\mathrm{d}\mathbf{y}_{1:\kappa}\,\mathrm{d}\mathbf{y}_{\kappa+1:n}$$
$$=\frac{\binom{n}{\kappa}}{\binom{n}{\kappa}c^\kappa(1-c)^{n-\kappa}}\int_{C^\kappa}g(\mathbf{y}_{1:\kappa})\,\mathrm{US}(\mathbf{y}_{1:\kappa})^2\,\mathrm{d}\mathbf{y}_{1:\kappa}\cdots$$
$$\underbrace{\int_{(G\setminus C)^{n-\kappa}}g(\mathbf{y}_{\kappa+1:n})\,\mathrm{d}\mathbf{y}_{\kappa+1:n}}_{=(1-c)^{n-\kappa}}$$
$$=\frac{\binom{n}{\kappa}(1-c)^{n-k}}{\binom{n}{\kappa}c^\kappa(1-c)^{n-\kappa}}\int_{C^\kappa}g(\mathbf{y}_{1:\kappa})\left(\frac{c}{\kappa}\sum_{i=1}^\kappa\frac{f(y_i)}{g(y_i)}h(y_i)\right)^2\mathrm{d}\mathbf{y}_{1:\kappa}$$
$$=\frac{c^2}{c^\kappa}\int_{C^\kappa}g(\mathbf{y}_{1:\kappa})\left(\frac{1}{\kappa}\sum_{i=1}^\kappa\frac{f(y_i)}{g(y_i)}h(y_i)\right)^2\mathrm{d}\mathbf{y}_{1:\kappa}$$
$$\stackrel{(c)}{=}c^2\int_{C^\kappa}\frac{g(\mathbf{y}_{1:\kappa})}{\Pr(k(\mathbf{X}_\kappa)=\kappa)}\left(\frac{1}{\kappa}\sum_{i=1}^\kappa\frac{f(y_i)}{g(y_i)}h(y_i)\right)^2\mathrm{d}\mathbf{y}_{1:\kappa}$$
$$=c^2\mathbf{E}_g\left[\left(\frac{1}{\kappa}\sum_{i=1}^\kappa\frac{f(X_i)}{g(X_i)}h(X_i)\right)^2\bigg|\mathbf{X}_\kappa\in C^\kappa\right]$$
$$\stackrel{(d)}{=}c^2\left(\frac{1}{\kappa}v+\mathbf{E}\left[\frac{f(X)}{g(X)}h(X)\bigg|X\sim g,X\in C\right]^2\right)$$
$$=c^2\left(\frac{1}{\kappa}v+\left(\int_C\frac{g(x)}{c}\frac{f(x)}{g(x)}h(x)\,\mathrm{d}x\right)^2\right)=\frac{c^2}{\kappa}v+\theta^2,\quad (6)$$

where **(a)** comes from **1)** the fact that there are $\binom{n}{\kappa}$ ways of ordering $n$ elements such that $\kappa$ are in $C$ and $n-\kappa$ are in $G\setminus C$, and **2)** the fact that US does not depend on the order of its inputs, **(b)** comes from **1)** the property that US($\mathbf{y}$) does not change if additional samples are appended to $\mathbf{y}$ that are not in $C$ and **2)** the fact that $g(\mathbf{y})$ can be decomposed into $g(\mathbf{y}_{1:\kappa})g(\mathbf{y}_{\kappa+1:n})$ since it represents the joint probability density function for $n$ independent and identically distributed random variables, **(c)** comes from the fact that $\Pr(k(\mathbf{X}_\kappa)=\kappa)=c^\kappa$, and **(d)** comes from Property 2.

Combining (5) with (6) we have that
$$\mathrm{Var}_g(\mathrm{US}(\mathbf{X}_n)|k(\mathbf{X}_n)>0)$$
$$=\left(\sum_{\kappa=1}^n\frac{\Pr(k(\mathbf{X}_n)=\kappa)}{\Pr(k(\mathbf{X}_n)>0)}\left(\frac{c^2}{\kappa}v+\theta^2\right)\right)-\theta^2$$
$$=c^2v\left(\sum_{\kappa=1}^n\frac{\Pr(k(\mathbf{X}_n)=\kappa)}{\Pr(k(\mathbf{X}_n)>0)}\frac{1}{\kappa}\right)$$
$$+\theta^2\underbrace{\left(\sum_{\kappa=1}^n\frac{\Pr(k(\mathbf{X}_n)=\kappa)}{\Pr(k(\mathbf{X}_n)>0)}\right)}_{=1}-\theta^2$$
$$=c^2v\sum_{\kappa=1}^n\frac{\Pr(k(\mathbf{X}_n)=\kappa)}{\Pr(k(\mathbf{X}_n)>0)}\frac{1}{\kappa}$$
$$=c^2v\mathbf{E}_{B(n,c)}\left[\frac{1}{\kappa}\bigg|\kappa>0\right].\qquad\blacksquare$$

**Theorem 9.** *If $F\cap H\subseteq G$ then*
$$\mathrm{Var}_g(\mathrm{IS}(\mathbf{X}_n)|k(\mathbf{X}_n>0))=v\frac{c}{n\rho}+\theta^2\frac{c\rho(n-1)+\rho-cn}{cn\rho^2}.$$

*Proof.* At a high level, this proof is similar to the proof of Theorem 8, but uses the property that $\mathrm{IS}(\mathbf{X}_n)=\frac{k(\mathbf{X}_n)}{cn}\mathrm{US}(\mathbf{X}_n)$.

$$\mathrm{Var}_g(\mathrm{IS}(\mathbf{X}_n)|k(\mathbf{X}_n)>0)$$
$$=\mathbf{E}_g[\mathrm{IS}(\mathbf{X}_n)^2|k(\mathbf{X}_n)>0]-\mathbf{E}_g[\mathrm{IS}(\mathbf{X}_n)|k(\mathbf{X}_n)>0]^2$$
$$\stackrel{(a)}{=}\mathbf{E}_g[\mathrm{IS}(\mathbf{X}_n)^2|k(\mathbf{X}_n)>0]-\left(\frac{\theta}{1-(1-c)^n}\right)^2$$
$$=\left(\sum_{\kappa=1}^n\frac{\Pr(k(\mathbf{X}_n)=\kappa)}{\Pr(k(\mathbf{X}_n)>0)}\mathbf{E}_g[\mathrm{IS}(\mathbf{X}_n)^2|k(\mathbf{X}_n)=\kappa]\right)$$
$$-\left(\frac{\theta}{1-(1-c)^n}\right)^2,\qquad(7)$$

where **(a)** comes from Theorem 6.

Also,
$$\mathbf{E}_g[\,\mathrm{IS}(\mathbf{X}_n)^2|k(\mathbf{X}_n)=\kappa]$$
$$\stackrel{(a)}{=}\mathbf{E}_g\left[\left(\frac{k(\mathbf{X}_n)}{cn}\mathrm{US}(\mathbf{X}_n)\right)^2\bigg|k(\mathbf{X}_n)=\kappa\right]$$
$$=\frac{\kappa^2}{c^2n^2}\mathbf{E}_g[\mathrm{US}(\mathbf{X}_n)^2|k(\mathbf{X}_n)=\kappa]\stackrel{(b)}{=}\frac{\kappa^2}{c^2n^2}\left(\frac{c^2}{\kappa}v+\theta^2\right),(8)$$

where **(a)** holds because $\mathrm{IS}(\mathbf{X}_n)=\frac{k(\mathbf{X}_n)}{cn}\mathrm{US}(\mathbf{X}_n)$ and **(b)** follows from (6). Using the shorthand, $\rho:=\Pr(k(\mathbf{X}_n)>0)=1-(1-c)^n$ and by combining (7) with (8) we have

that:

$$\text{Var}_g(\text{IS}(\mathbf{X}_n)|k(\mathbf{X}_n) > 0)$$
$$= \left(\sum_{\kappa=1}^n \frac{\Pr(k(\mathbf{X}_n) = \kappa)}{\Pr(k(\mathbf{X}_n) > 0)} \frac{\kappa^2}{c^2 n^2} \left(\frac{c^2}{\kappa}v + \theta^2\right)\right)$$
$$\quad - \left(\frac{\theta}{1-(1-c)^n}\right)^2$$
$$= \frac{v}{n^2\rho}\underbrace{\left(\sum_{\kappa=1}^n \Pr(k(\mathbf{X}_n) = \kappa)\kappa\right)}_{=\mathbf{E}_{B(n,c)}[\kappa]=nc}$$
$$\quad + \frac{\theta^2}{c^2 n^2 \rho}\underbrace{\left(\sum_{\kappa=1}^n \Pr(k(\mathbf{X}_n) = \kappa)\kappa^2\right)}_{=\mathbf{E}_{B(n,c)}[\kappa^2]=nc((n-1)c+1)} - \left(\frac{\theta}{\rho}\right)^2$$
$$= v\frac{c}{n\rho} + \frac{\theta^2((n-1)c+1)}{cn\rho} - \frac{\theta^2}{\rho^2}$$
$$= v\frac{c}{n\rho} + \theta^2 \frac{c\rho(n-1)+\rho-cn}{cn\rho^2}. \qquad \blacksquare$$

**Theorem 10.** *If $F \cap H \subseteq G$ then*
$$\text{Var}_g(\text{US}(\mathbf{X}_n)) = \rho c^2 v \mathbf{E}_{B(n,c)}\left[\frac{1}{\kappa}\bigg|\kappa > 0\right] + \theta^2\rho(1-\rho).$$

*Proof.*
$$\text{Var}_g(\text{US}(\mathbf{X}_n)) = \mathbf{E}_g[\text{US}(\mathbf{X}_n)^2] - \mathbf{E}_g[\text{US}(\mathbf{X}_n)]^2$$
$$\stackrel{(a)}{=} \mathbf{E}_g[\text{US}(\mathbf{X}_n)^2] - \rho^2\theta^2$$
$$= \left(\sum_{\kappa=0}^n \Pr(k(\mathbf{X}_n) = \kappa)\mathbf{E}_g[\text{US}(\mathbf{X}_n)^2|k(\mathbf{X}_n) = \kappa]\right)$$
$$\quad - \rho^2\theta^2$$
$$= \Pr(k(\mathbf{X}_n) = 0)\underbrace{\mathbf{E}_g[\text{US}(\mathbf{X}_n)^2|k(\mathbf{X}_n) = 0]}_{=0}$$
$$\quad + \left(\sum_{\kappa=1}^n \Pr(k(\mathbf{X}_n) = \kappa)\mathbf{E}_g[\text{US}(\mathbf{X}_n)^2|k(\mathbf{X}_n) = \kappa]\right)$$
$$\quad - \rho^2\theta^2$$
$$\stackrel{(b)}{=} \rho\left(\sum_{\kappa=1}^n \frac{\Pr(k(\mathbf{X}_n) = \kappa)}{\rho}\left(\frac{c^2}{\kappa}v + \theta^2\right)\right) - \rho^2\theta^2$$
$$= \rho c^2 v \left(\sum_{\kappa=1}^n \frac{\Pr(k(\mathbf{X}_n) = \kappa)}{\rho}\frac{1}{\kappa}\right)$$
$$\quad + \rho\theta^2 \underbrace{\left(\sum_{\kappa=1}^n \frac{\Pr(k(\mathbf{X}_n) = \kappa)}{\rho}\right)}_{=1} - \rho^2\theta^2$$
$$= \rho c^2 v \mathbf{E}_{B(n,c)}\left[\frac{1}{\kappa}\bigg|\kappa > 0\right] + \theta^2\rho(1-\rho),$$

where **(a)** comes from Theorem 7, **(b)** comes from (6) and from multiplying one term by $\rho/\rho = 1$. $\blacksquare$

**Theorem 11.** *If $F \cap H \subseteq G$ then*
$$\text{Var}_g(\text{IS}(\mathbf{X}_n)) = \frac{1}{n}\left(cv + \theta^2\left(\frac{1}{c}-1\right)\right).$$

*Proof.*
$$\text{Var}_g(\text{IS}(\mathbf{X}_n)) \stackrel{(a)}{=} \frac{1}{n}\text{Var}_g(\text{IS}(X))$$
$$= \frac{1}{n}\left(\mathbf{E}_g[\text{IS}(X)^2] - \mathbf{E}_g[\text{IS}(X)]^2\right)$$
$$\stackrel{(b)}{=} \frac{1}{n}\left(\mathbf{E}_g[\text{IS}(X)^2] - \theta^2\right)$$
$$= \frac{1}{n}\bigg(\Pr(X \in C|X \sim g)\mathbf{E}_g[\text{IS}(X)^2|X \in C]$$
$$\quad + \Pr(X \notin C|X \sim g)\underbrace{\mathbf{E}_g[\text{IS}(X)^2|X \notin C]}_{=0} - \theta^2\bigg)$$
$$= \frac{1}{n}\left(c\mathbf{E}_g[\text{IS}(X)^2|X \in C] - \theta^2\right)$$
$$\stackrel{(c)}{=} \frac{1}{n}\left(c\left(v + \frac{\theta^2}{c^2}\right) - \theta^2\right)$$
$$= \frac{1}{n}\left(cv + \theta^2\left(\frac{1}{c}-1\right)\right),$$

where **(a)** holds because $\text{IS}(\mathbf{X}_n)$ is the sum of $n$ independent and identically distributed random variables, **(b)** comes from Property 1, and **(c)** comes from applying (8) with $n = 1$ and $\kappa = 1$. $\blacksquare$

**Property 3.** $c\rho(n-1) + \rho - cn \geq 0$,

*Proof.* Recall that $\rho := 1 - (1-c)^n$, so we have that:
$$c\rho(n-1) + \rho - cn = c(1-(1-c)^n)(n-1) + 1 - (1-c)^n - cn$$
$$= (cn-c)(1-(1-c)^n) + 1 - (1-c)^n - cn$$
$$= cn - cn(1-c)^n - c + c(1-c)^n + 1 - (1-c)^n - cn$$
$$= (1-c)^n(-cn+c-1) - c + 1. \qquad (9)$$

We will show by induction that (9) is non-negative for all $n \geq 1$. First, notice that for the base case where $n = 1$, (9) is equal to zero. For the inductive step we will show that (9) is non-negative for $n+1$ given that it is non-negative for $n$.

$$(1-c)^{n+1}(-c(n+1)+c-1) - c + 1$$
$$= (1-c)(1-c)^n(-cn+c-1) - (1-c)^{n+1}c$$
$$\quad + (-c+1)(1-c+c)$$
$$= (1-c)\underbrace{\left((1-c)^n(-cn+c-1) - c + 1\right)}_{(a)}$$
$$\quad - (1-c)^{n+1}c + c(1-c),$$

where **(a)** is positive by the inductive hypothesis, and so we need only show that $-(1-c)^{n+1}c + c(1-c) \geq 0$. Since
$$-(1-c)^{n+1}c + c(1-c) = c\Big((1-c) - (1-c)^{n+1}\Big),$$
and $1-c \geq (1-c)^{n+1}$ because $c \in (0,1]$, we conclude. $\blacksquare$